%% file: ijcai22.tex
\documentclass{article}
\usepackage{ijcai22}

\usepackage{times}
\usepackage{soul}
\usepackage{url}
\usepackage[hidelinks]{hyperref}
\usepackage[utf8]{inputenc}
\usepackage[small]{caption}
\usepackage{graphicx}
\usepackage{amsmath}
\usepackage{amsthm}
\usepackage{booktabs}
\usepackage{algorithm}
\usepackage{algorithmic}
\usepackage{multirow}
\usepackage{subcaption}
\usepackage{amsmath}
\usepackage{amssymb}
\usepackage{makecell}

\newcommand{\citet}[1]{\citeauthor{#1}~\shortcite{#1}}

\title{Efficient Document-level Event Extraction via Pseudo-Trigger-aware Pruned Complete Graph}

\author{
    Tong Zhu$^1$ \and
    Xiaoye Qu$^2$ \and
    Wenliang Chen$^1$\thanks{Corresponding author} \and
    Zhefeng Wang$^2$ \and \\
    Baoxing Huai$^2$ \and
    Nicholas Yuan$^2$ \And
    Min Zhang$^1$
    \affiliations
    $^1$Institute of Artificial Intelligence, School of Computer Science and Technology, \\Soochow University, China\\
    $^2$Huawei Cloud, China
    \emails
    tzhu7@stu.suda.edu.cn, \{wlchen,minzhang\}@suda.edu.cn\\
    \{quxiaoye,wangzhefeng,huaibaoxing\}@huawei.com, nicholas.jing.yuan@gmail.com
}

\begin{document}

\maketitle

\input{abstract}
\input{introduction}
\input{methodology}
\input{experiments}
\input{related_work}
\input{conclusion}
\input{acknowledgement}

%% The file named.bst is a bibliography style file for BibTeX 0.99c
\bibliographystyle{named}
\bibliography{ijcai22}

\end{document}

%% file: abstract.tex
\begin{abstract}
    Most previous studies of document-level event extraction mainly focus on building argument chains in an autoregressive way, which achieves a certain success but is inefficient in both training and inference.
    In contrast to the previous studies, we propose a fast and lightweight model named as PTPCG.
    In our model, we design a novel strategy for event argument combination together with a non-autoregressive decoding algorithm via pruned complete graphs, which are constructed under the guidance of the automatically selected pseudo triggers.
    Compared to the previous systems, our system achieves competitive results with 19.8\% of parameters and much lower resource consumption, taking only 3.8\% GPU hours for training and up to 8.5 times faster for inference.
    Besides, our model shows superior compatibility for the datasets with (or without) triggers and the pseudo triggers can be the supplements for annotated triggers to make further improvements.
    Codes are available at \url{https://github.com/Spico197/DocEE} .
\end{abstract}

%% file: introduction.tex
\section{Introduction}

\begin{figure}[t]
    \centering
    \includegraphics[width=\linewidth]{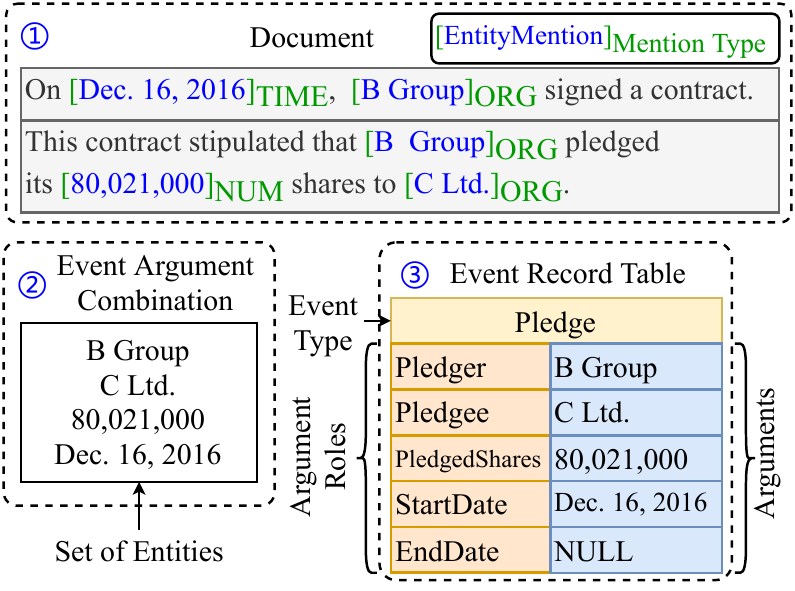}
    \caption{Examples of event extraction.}
    \label{fig:dee_example}
\end{figure}

Event Extraction (EE) aims at filling event tables with given texts.
Different from sentence-level EE (SEE) which focuses on building trigger-centered trees~\cite{chen-etal-2015-dmcnn,nguyen-etal-2016-jrnn,liu-etal-2018-jmee,wadden-etal-2019-dygiepp,lin-etal-2020-oneie}, document-level EE (DEE) is to decode argument combinations from abundant entities across multiple sentences and fill these combinations into event record tables as shown in Figure~\ref{fig:dee_example}, where annotated triggers are often not available~\cite{yang-etal-2018-dcfee,zheng-etal-2019-doc2edag,xu-etal-2021-git}.

One of the challenges for DEE is event argument combination without triggers.
Triggers in DEE datasets are either absent or in a low annotation quality since the large-scale datasets are usually generated via distantly supervised (DS) alignment with existing knowledge bases (KB)~\cite{mintz-2009-ds,chen-etal-2017-dsdata}.
Consequently, the absence of triggers drives the development of trigger-free argument combination methods in recent studies.
\citet{yang-etal-2018-dcfee} first identify a key sentence and then fill the event record table by finding arguments near the sentence, while partial global features and arguments are still missing.
\citet{zheng-etal-2019-doc2edag} and \citet{xu-etal-2021-git} fully utilize the global context and make significant improvements by building an directed acyclic graph (DAG).
However, such DAG-based methods require massive computing resources as they rely on an autoregressive fashion to decode argument combinations, which is inefficient in both training and inference for long documents. Meanwhile, building DAGs consumes lots of memories to store previous paths.
As a result, it takes \textbf{almost one week with a minimum of four 32GB GPUs to train such DAG-based DEE models, and the inference speed is also extremely slow.}

Considering the above challenge in DEE and the speed \& memory consuming problems in the DAG-based methods, in this paper, we aim to 1) propose a universal event argument combination strategy that works on both trigger-aware and trigger-free DEE; and 2) provide a blazing fast and lightweight model for document-level event extraction.
We propose a novel non-autoregressive approach named as \textbf{P}seudo-\textbf{T}rigger-aware \textbf{P}runed \textbf{C}omplete \textbf{G}raph (PTPCG).
Specifically, we formulate each argument combination as a pruned complete graph, where the important arguments are identified and treated as a group of pseudo triggers with bidirectional connections to each other and other ordinary arguments are linked from these pseudo triggers in a directed manner.
Based on the pruned complete graph with the pseudo triggers, we design an efficient algorithm with the non-autoregressive decoding strategy for event argument combination extraction.

The experiments results show that our PTPCG can reach competitive results with only 19.8\% parameters of DAG-based SOTA models, just taking 3.8\% GPU hours to train and up to 8.5 times faster in terms of inference.
Besides, our PTPCG is highly flexible and scalable that can be used as a general architecture for non-autoregressive trigger-based event extraction.
If only one pseudo trigger is selected for each combination, the pruned complete graph becomes a trigger-centered tree like SEE.
Furthermore, pseudo triggers can be adopted as supplements to enhance the annotated-trigger-based methods.

In summary, our contributions include:
\begin{itemize}
    \item We propose a novel non-autoregressive event argument combination paradigm based on pruned complete graph with pseudo triggers, which is compatible in document-level event extraction with (or without) triggers.
    \item Our model is fast and lightweight for end-to-end document-level event extraction and we conduct extensive experiments to show the efficiency and efficacy.
    \item To our best knowledge, our present approach is the first work that explores the effects of using some arguments as pseudo triggers in DEE, and we design a metric to help select a group of pseudo triggers automatically. Furthermore, such metric can also be used for measuring the quality of annotated triggers in DEE.
\end{itemize}

%% file: methodology.tex
\section{Methodology}

\begin{figure*}[t]
    \centering
        \includegraphics[width=\linewidth]{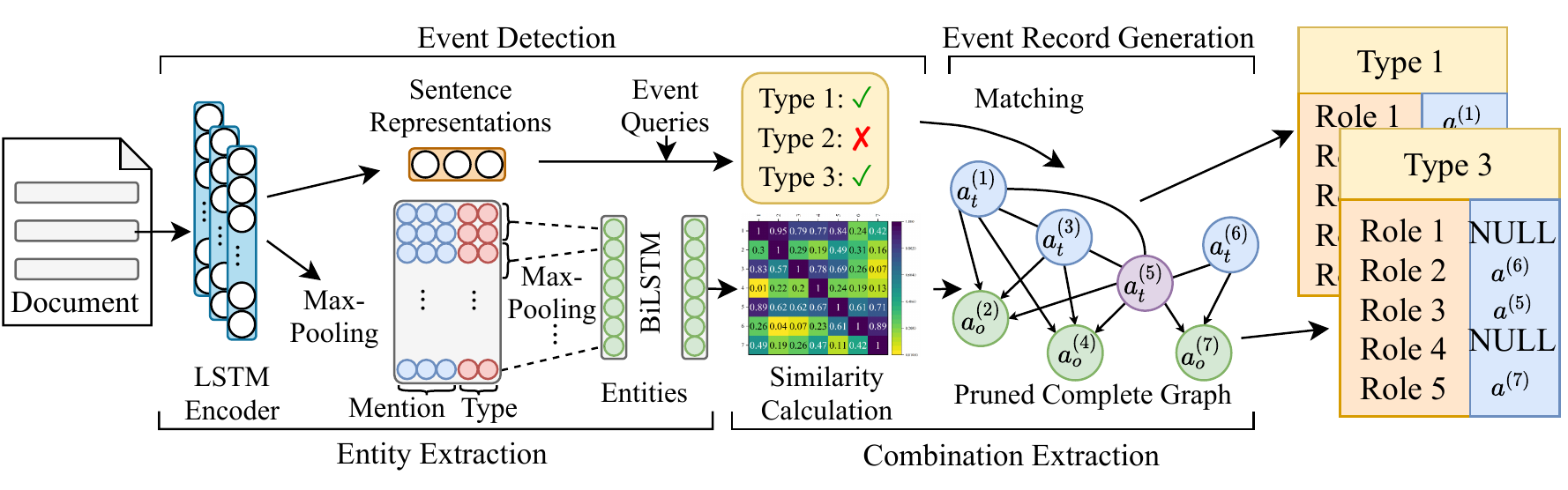}
    \caption{Overview of our PTPCG. Event types and entity mentions are first extracted with a shared LSTM encoder, then similarities between entity pairs are calculated to help recover the adjacent matrix of pruned complete graph. After combinations are decoded from the adjacent matrix, the extracted types are paired with combinations to generate final event records.}
    \label{fig:framework}
\end{figure*}

As shown in Figure~\ref{fig:framework}, our model can be divided into four components:
1) \textbf{Event detection} performs multi-label classification to identify all possible event types.
2) \textbf{Entity extraction} extracts all entities from the documents and encodes these entities into dense vectors.
3) \textbf{Combination extraction} builds pruned complete graphs and decodes argument combinations from such graphs.
4) \textbf{Event record generation} combines the results of event types and extracted argument combinations to generate the final event records.

\subsection{Event Detection}

For a document $\mathcal{D}$, a bidirectional long short-term memory (BiLSTM) network~\cite{hochreiter1997lstm} is used to encode each sentence $s_i$ into token-wise hidden states $(h_i^{(1)}, h_i^{(2)}, \ldots, h_i^{(|s_i|)})$, where $h_i^{(j)} \in \mathbb{R}^{d_h}$ is the concatenation of two direction representations $\overrightarrow{ h_i^{(j)} } \| \overleftarrow{ h_i^{(j)} }$ and $|s_i|$ is the length of $i$-th sentence.
The last hidden states in each direction are concatenated to get the sentence representation $g_i \in \mathcal{G}$ as $g_i = \overrightarrow{ h_i^{(|s_i|)} } \| \overleftarrow{ h_i^{(0)} }$.

Subsequently, we follow Doc2EDAG~\cite{zheng-etal-2019-doc2edag} and use randomly initialized event queries with multi-head attention on $\mathcal{G}$ to make a binary classification for each event type.
The loss function of event detection component $\mathcal{L}_{\text{det}}$ is defined as a binary cross entropy loss.

\subsection{Entity Extraction}

The entity extraction task can be formulated as a sequence tagging task in BIO scheme~\cite{zheng-etal-2019-doc2edag}.
To enhance the entity recognition, we add more \textit{money}, \textit{date}, \textit{percentage ratio} and \textit{shares} entities into the dataset by simple regular expression matching.
To deal with the document-level entity extraction, we first split the whole document into sentences and perform a sentence-level mention extraction.
Then, we use BiLSTM with CRF~\cite{lample-etal-2016-bilstmcrf} to extract entity mentions with the same BiLSTM used in event detection.
The training objective of entity extraction is to minimize the negative log-likelihood loss $\mathcal{L}_{\text{ent}}$ of CRF for each sentence.

For all tokens of a mention, a max-pooling operation is applied on token-level representations to get the mention representation $\tilde{m}_j$.
As mention types have been proved effective for downstream sub-modules~\cite{zheng-etal-2019-doc2edag,xu-etal-2021-git}, we convert the predicted discrete mention types into vectors by looking up an embedding table.
After concatenating $\tilde{m}_j$ and type embeddings $l_j$, we get the final mention representation $m_j = \tilde{m}_j \| l_j \in \mathbb{R}^{d_a}$, where $d_a = d_h + d_l$ and $d_l$ denotes the dimension of $l_j$.
Finally, all the mentions for an entity are aggregated to get the entity representation $\hat{e}_i$ via another max-pooling.
For better modeling entity semantics and the latent connections for combiantion extraction, an additional BiLSTM layer is applied to get the set of entity representations $\mathcal{E} = \{\tilde{e}_i\}_{i=1}^{|\mathcal{E}|}$.

\subsection{Combination Extraction}

In this section, we introduce the details of selecting pseudo triggers, building pruned complete graphs, and decoding combinations from the graphs.

\subsubsection{Pseudo Trigger Selection}
It is hard to annotate conventional triggers manually in documents due to the document length and dataset scale.
We instead select a group of pseudo triggers for each event type in an automatic way.

Empirically, the triggers are keywords that play two roles: 1) triggers can be used to identify combinations; 2) triggers are fingerprints that can distinguish different combinations.
Combinations are made of arguments and we can extract a specific combination by finding all the corresponding arguments.
To this end, we design an importance metric that evaluates the possibility of whether a group of arguments can serve as the pseudo triggers.
In general, we first select a group of argument roles as candidates for each event type according to the importance scores, and take the corresponding arguments as the pseudo triggers.

Formally, the importance score is obtained by the existence and the distinguishability.
For a subset of predefined argument roles in type $t_i$, $\mathcal{R} = \{r_j\}_{j=1}^{|\mathcal{R}|}$ are selected as the pseudo trigger candidates.
$|\mathcal{R}|$ is a hyper-parameter that denotes the number of selected pseudo triggers for each argument combination.
The existence measures whether the arguments of $\mathcal{R}$ can identify combinations.
$N_e^{(\mathcal{R})}$ is the number of event records that at least one corresponding argument of $\mathcal{R}$ is not NULL, and $N^{(i)}$ is the number of total records of $t_i$.
The distinguishability is defined to satisfy that the triggers can distinguish different combinations, where $N_u^{(\mathcal{R})}$ is the number of records that the arguments of $\mathcal{R}$ do not appear in other records in the same document.
With the multiplication of existence and distinguishability, pseudo triggers are selected by picking the candidate with the highest importance score.

\begin{equation}
    \label{eqn:importance}
    \begin{gathered}
        \textnormal{Existence}(\mathcal{R}) = \frac{ N_e^{(\mathcal{R})} }{N^{(i)}}, \quad\textnormal{Distinguish}(\mathcal{R}) = \frac{ N_u^{(\mathcal{R})} }{N^{(i)}} \\
        \textnormal{Importance}(\mathcal{R}) = 
        \textnormal{Existence}(\mathcal{R}) \times \textnormal{Distinguish}(\mathcal{R})
    \end{gathered}
\end{equation}

\subsubsection{Pruned Complete Graph Construction}

Based on the DEE task setting and data analysis, we propose an assumption that \textit{arguments in the same combination are close to each other in the semantic space}.
Following this assumption, we take the pseudo triggers as the core of argument combinations and formulate each combination as a pruned complete graph.
As shown in the pruned complete graph of Figure~\ref{fig:framework}, for any two arguments as pseudo triggers $a^{(i)}_t$ and $a^{(j)}_t$ in the same combination, they are bidirectionally connected, where the adjacent matrix $y^{(i,j)}_A = y^{(j,i)}_A = 1$.
For a pseudo trigger $a^{(i)}_t$ and an ordinary argument $a^{(j)}_o$ in the same combination, they are connected with a directional link and $y^{(i,j)}_A = 1$.
Besides, each argument $a^{(i)}$ has a self-loop connection where $y^{(i,i)}_A = 1$.
Other entries in $y_A$ are zeros, where entities that do not participate in any combination are isolated nodes in the graph.

After obtaining entity representations, a dot scaled similarity function (Eqn~\ref{eqn:sim_calc}) is applied to estimate their semantic distances:
\begin{equation}
    \label{eqn:sim_calc}
    \begin{gathered}
    \tilde{e_i} = e_i \times W_s^{\top} + b_s, \quad \tilde{e_j} = e_j \times W_e^{\top} + b_e \\
    \tilde{A}_{i,j} = \textnormal{sigmoid}\left(\tilde{e_i}^{\top}  \tilde{e_j} / \sqrt{d_h}\right)
    \end{gathered}
\end{equation}
where $\tilde{A}$ denotes the similarity matrix, $W_s, W_e \in \mathbb{R}^{d_a \times d_a}$ and $b_s, b_e \in \mathbb{R}^{d_a}$ are trainable parameters for semantic space linear projection.

In training, we use binary cross entropy function to formulate the combination loss $\mathcal{L}_{\text{comb}}$.

To predict the binary adjacent matrix $A$ of the pruned complete graph for further decoding, the threshold $\gamma$ is used here (Eqn~\ref{eqn:adjacent}).

\begin{equation}
    \label{eqn:adjacent}
    A_{i,j} = \begin{cases}
        1 & \tilde{A}_{i,j} \geqslant \gamma \\
        0 & otherwise
    \end{cases}
\end{equation}

\subsubsection{Non-Autoregressive Combination Decoding}

\begin{figure}
    \centering
    \includegraphics[width=\linewidth]{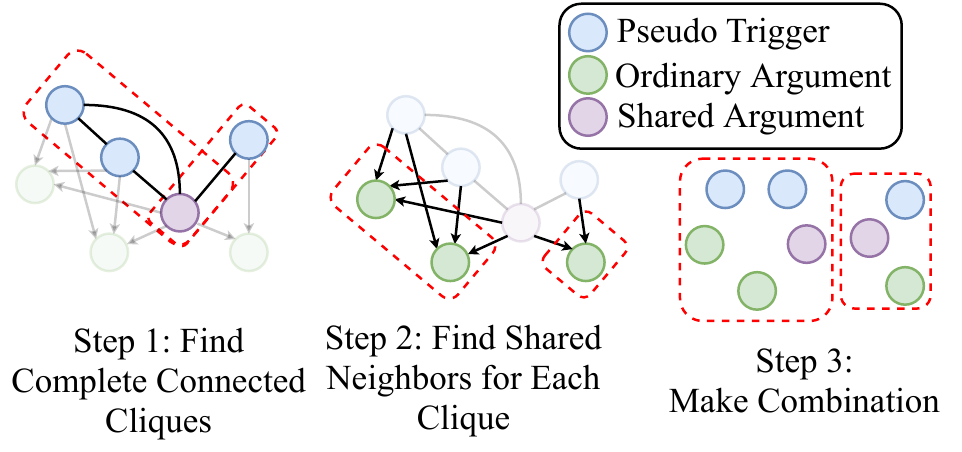}
    \caption{Combination decoding (best viewed in color).}
    \label{fig:combination_decoding}
\end{figure}

Event argument combinations are extracted based on the predicted adjacent matrix $A$ with a non-autoregressive decoding algorithm.

First, all the pseudo triggers are identified based on nodes' out-degree and each pseudo trigger group is recognized as a clique.
If the out-degree of an entity is greater than 0 except for self-loop, then the entity is treated as a pseudo trigger.
For $|\mathcal{R}|$ = 1, all the combinations are pseudo-trigger-centered trees, where each combination is made of a pseudo trigger with its neighbors.
Otherwise, Bron-Kerbosch (BK) algorithm~\cite{bron-kerbosch} is applied first to find all possible cliques (step 1 in Figure~\ref{fig:combination_decoding}).
To apply BK algorithm, the links between arguments must be undirected, thus we first extract all bidirectional links as undirected input.

As shown in Figure~\ref{fig:combination_decoding}, the next step is to find the ordinary (non-pseudo-trigger) arguments in the combination.
We further exploit all the neighbors of each pseudo trigger in the clique.
After that, an intersection operation is performed to find commonly shared ordinary arguments.
The combination is consist of a pseudo trigger clique and their commonly shared ordinary arguments.
For those extreme records that have only one argument in the combinations, all predicted entities are aggregated together as a default combination.

The document length does not affect the event record generation speed, and the time complexity of our event records generation is polynomial ($\mathcal{O}(N_t \times N_c)$), while DAG-based model is $\mathcal{O}(N_t \times N_s \times N_r)$, where $N_t$, $N_c$, $N_s$, $N_r$ are the number of predicted types, combinations, predicted spans and the average number of roles per type.
In most cases, $N_s \times N_r \gg N_c$, so our PTPCG consistently holds the speed advantage.

\subsection{Event Records Generation}

After the set of combinations $\mathcal{C}$ are obtained from the pruned complete graphs, the next step is to fill these combinations into event tables.
First, all the combinations should match with event types.
Since event detection is a multi-label classification task, there may be more than one type prediction.
For all type predictions $\mathcal{T}_p = \{t_j\}_{j=1}^{|\mathcal{T}_p|}$ and combinations $\mathcal{C}$, we perform a Cartesian product and get all type-combination pairs $\{ <t_j, c_k> | 1 \leqslant j \leqslant |\mathcal{T}_p|, 1 \leqslant k \leqslant |\mathcal{C}| \}$.

For each pair $<t_j, c_k>$, we use an event-relevant feed-forward network (FFN) as classifier to get possible role results for all arguments $\mathcal{E}_k$ in $c_k$, and use sigmoid function to get the probabilities $p^{(j)}_{role}(r_j|c_k)$.
The loss $\mathcal{L}_{\text{role}}$ is calculated by a binary cross entropy function.

In the role classification task, an entity can act as more than one role in a record, while a role in a table can only be filled with one entity.
Following this setting, we take the entity $e_{i^*}^{(k)}$ as the argument of role $r_j^{(q)}$ if and only if it satisfies the constraint in Eqn~\ref{eqn:role_pred}.
If $\forall{j} \in [1, |\mathcal{T}_p|], p^{(j)}_{role}(r_j|c_k) < 0.5$, then the pair is recognized as an invalid candidate and will be dropped.

\begin{equation}
    i^* = \textnormal{argmax}_q p^{(j)}_{role}(r_j^{(q)}|c_k)
    \label{eqn:role_pred}
\end{equation}

\subsection{Optimization}

Our PTPCG is an end-to-end model with joint training and scheduled sampling~\cite{bengio2015scheduledsampling} strategy.
The overall loss is a weighted sum of all losses as below:

\begin{equation}
    \mathcal{L} = \alpha_1 \mathcal{L}_{\text{det}} + \alpha_2 \mathcal{L}_{\text{ent}} + \alpha_3 \mathcal{L}_{\text{comb}} + \alpha_4 \mathcal{L}_{\text{role}}
    \label{eqn:total_loss}
\end{equation}
where $\alpha_1, \alpha_2, \alpha_3, \alpha_4$ are hyper-parameters to reduce the unbalanced loss effect.

%% file: experiments.tex
\section{Experiments}
\subsection{Datasets}

We use ChFinAnn \cite{zheng-etal-2019-doc2edag} and DuEE-fin \cite{duee-fin} datasets to make fair comparisons across all methods.

\noindent (1) \textbf{ChFinAnn} is by far the largest DEE dataset constructed by distantly supervised alignment \textbf{without trigger annotations}, which is widely used in the previous studies.
This dataset contains 32k financial announcements and 48k event records, in which 29\% documents have more than one record and 98\% of records have arguments scattered across different sentences.
On average a document contains 20 sentences and the longest document contains 6.2k Chinese characters.
(2) \textbf{DuEE-fin} is another dataset for DEE \textbf{with trigger annotations}.
It contains 13 event types and 11.7k documents, where the test set is evaluated online.
Each record in DuEE-fin has a labeled trigger word without specific positions, and 36\% of records share the same trigger in a document.

\subsection{Experiment Settings}
We choose the hyper-parameters of our system according to the performance on the development set in ChFinAnn.
In PTPCG, we use 2 layers of shared BiLSTM for event detection and entity extraction, and another 2 layers of BiLSTM for entity encoding.
We use the same vocabulary as \cite{zheng-etal-2019-doc2edag} and randomly initialize all the embeddings where $d_h$=768 and $d_l$=32.
Adam \cite{adam} optimizer is used with a learning rate of 5e-4 and the mini-batch size is 64.
The weights in Equation~\ref{eqn:total_loss} are 0.05, 1.0, 1.0, and 1.0, and $\gamma$ in Equation~\ref{eqn:adjacent} is 0.5.
Following the setting in \citet{zheng-etal-2019-doc2edag}, we train our models for 100 epochs and select the checkpoint with the best F1 score on the development set to evaluate on the test set.

\begin{table*}[tb]
    \centering
    \small
    \begin{tabular}{lcrcccccccccccc}
        \toprule
        \multirow{2}{*}{Model} & \#Params & GPU & \multicolumn{3}{c}{ChFinAnn-Single} & \multicolumn{3}{c}{ChFinAnn-All} & \multicolumn{3}{c}{DuEE-fin w/o Tgg} & \multicolumn{3}{c}{DuEE-fin w/ Tgg}\\
        \cmidrule(lr){4-6}\cmidrule(lr){7-9}\cmidrule(lr){10-12}\cmidrule(lr){13-15}
        &(w/o Emb) & Hours & P & R & F1 & P & R & F1 & P & R & F1 & P & R & F1 \\
        \midrule
        DCFEE-O$^{*}$ & 32M (16M) & 192.0 & 73.2 & 71.6 & 72.4 & 69.7 & 57.8 & 63.2 & 56.2 & 48.2 & 51.9 & 51.9 & 49.6 & 50.7 \\
        DCFEE-M$^{*}$ & 32M (16M) & 192.0 & 64.9 & 71.7 & 68.1 & 60.1 & 61.3 & 60.7 & 38.7 & 52.3 & 44.5 & 37.3 & 48.6 & 42.2 \\
        GreedyDec$^{*}$ & 64M (48M) & 604.8 & 83.9 & 77.3 & 80.4 & 81.9 & 51.2 & 63.0 & 59.6 & 41.8 & 49.1 & 59.0 & 42.1 & 49.2 \\
        Doc2EDAG$^{*}$ & 64M (48M) & 604.8 & 83.2 & 89.3 & 86.2 & 81.1 & 77.0 & 79.0 & 66.7 & 50.0 & 57.2 & 67.1 & 51.3 & \textbf{58.1} \\
        GIT$^{*}$ & 97M (81M) & 633.6 & 85.0 & 88.7 & 86.8 & 82.4 & \textbf{77.6} & \textbf{79.9} & \textbf{68.2} & 43.4 & 53.1 & \textbf{70.3} & 46.0 & 55.6 \\
        \midrule
        PTPCG$_{|\mathcal{R}|=1}$ & 32M (16M) & \textbf{24.0} & \textbf{86.3} & \textbf{90.1} & \textbf{88.2} & \textbf{83.7} & 75.4 & 79.4 & 64.5 & \textbf{56.6} & \textbf{60.3} & 63.6 & \textbf{53.4} & \textbf{58.1} \\
        \bottomrule
    \end{tabular}
    \caption{
        Main results.
        \#Params is estimated on the ChFinAnn dataset.
        w/o Emb means the number of parameters without vocabulary embeddings.
        All models are trained with 100 epochs.
        -Single denotes the evaluation results on documents with only one event record.
        Tgg denotes the manually annotated triggers.
        $^{*}$We reproduce the results using their open-source codes.
    }
    \label{tab:exp_results_comparison}
\end{table*}

% \subsection{Baselines}
\subsection{Baselines and Metrics}

\paragraph{Baselines.}
(1) \textbf{DCFEE} \cite{yang-etal-2018-dcfee} has two variants here: DCFEE-O extracts only one record from one document while DCFEE-M extracts records as much as possible.
(2) \textbf{Doc2EDAG} \cite{zheng-etal-2019-doc2edag} constructs records as argument chains (DAG) and uses an auto-regressive way to extract final results.
(3) \textbf{GreedyDec} is a baseline from \citet{zheng-etal-2019-doc2edag} which fills one event table greedily.
(4) \textbf{GIT} \cite{xu-etal-2021-git} is another variant of Doc2EDAG and utilizes a graph neural network to further encode entities and add more features in DAG generation.

\paragraph{Metrics.}
We use the same evaluation setting as in the previous studies \cite{zheng-etal-2019-doc2edag,xu-etal-2021-git}.
For each predicted record, a golden record is selected without replacement by matching the record that has the same event type and the most shared arguments, and F1 scores are calculated by comparing arguments.
DuEE-fin uses an online evaluation fashion and the platform only provides micro-averaged scores.

\subsection{Main Results}

As shown in Table~\ref{tab:exp_results_comparison}, PTPCG achieves better or competitive results than the previous systems.
On the ChFinAnn dataset, our PTPCG achieves the best scores evluated on ChFinAnn-Single.
As for ChFinAnn-All, our model outperforms Doc2EDAG and is very close to GIT.
On the DuEE-fin dataset, PTPCG achieves the best F1 scores whether or not trigger words are provided, demonstrating good compatibility and universality.
Our PTPCG outperforms GIT and improves the overall F1 with 7.2\% w/o manually annotated triggers.
For DuEE-fin w/ Tgg, we find our approach declines from 60.3\% to 58.1\%.
After anaylzing the dataset, we find that the manually annotated triggers have less importance scores (62.9) than our pseudo triggers (83.8, $|\mathcal{R}|$=1).
Although adding annotated triggers when $|\mathcal{R}|$=1 can boost the importance score to 93.7, it is a trade-off between the number of triggers (annotated \& pseudo) and the adjacent matrix prediction.
We will discuss more in Section~\ref{sec:err_analyze}.

\subsection{Comparison on Model Parameters and Speed}
\paragraph{Comparison on the Number of Parameters.}
As shown in Table~\ref{tab:exp_results_comparison}, among all the models, PTPCG is the most lightweight one and is in the same scale with DCFEE,
while PTPCG outperforms DCFEE-O with 16.2\% absolute gain in F1 scores on ChFinAnn-All.
Without considering 16M vocabulary embedding parameters, PTPCG takes only 19.8\% parameters of GIT and reaches competitive results.

\paragraph{Comparison on Speed.}
Benefiting from the lightweight architecture design and non-autoregressive decoding style, our PTPCG is fast in both training and inference.

Comparing the training time as shown in Table~\ref{tab:exp_results_comparison}, it takes 633.6 GPU hours to train GIT, while PTPCG is 26.4 times faster and takes only 24.0 GPU hours.
These facts indicate that PTPCG is more efficient in training and requires much lower computation resource cost than other methods, and the whole training process has been reduced from almost a week with 4 NVIDIA V100 GPUs to just one day with 1 GPU.

\begin{figure}[t]
    \centering
    \includegraphics[width=\linewidth]{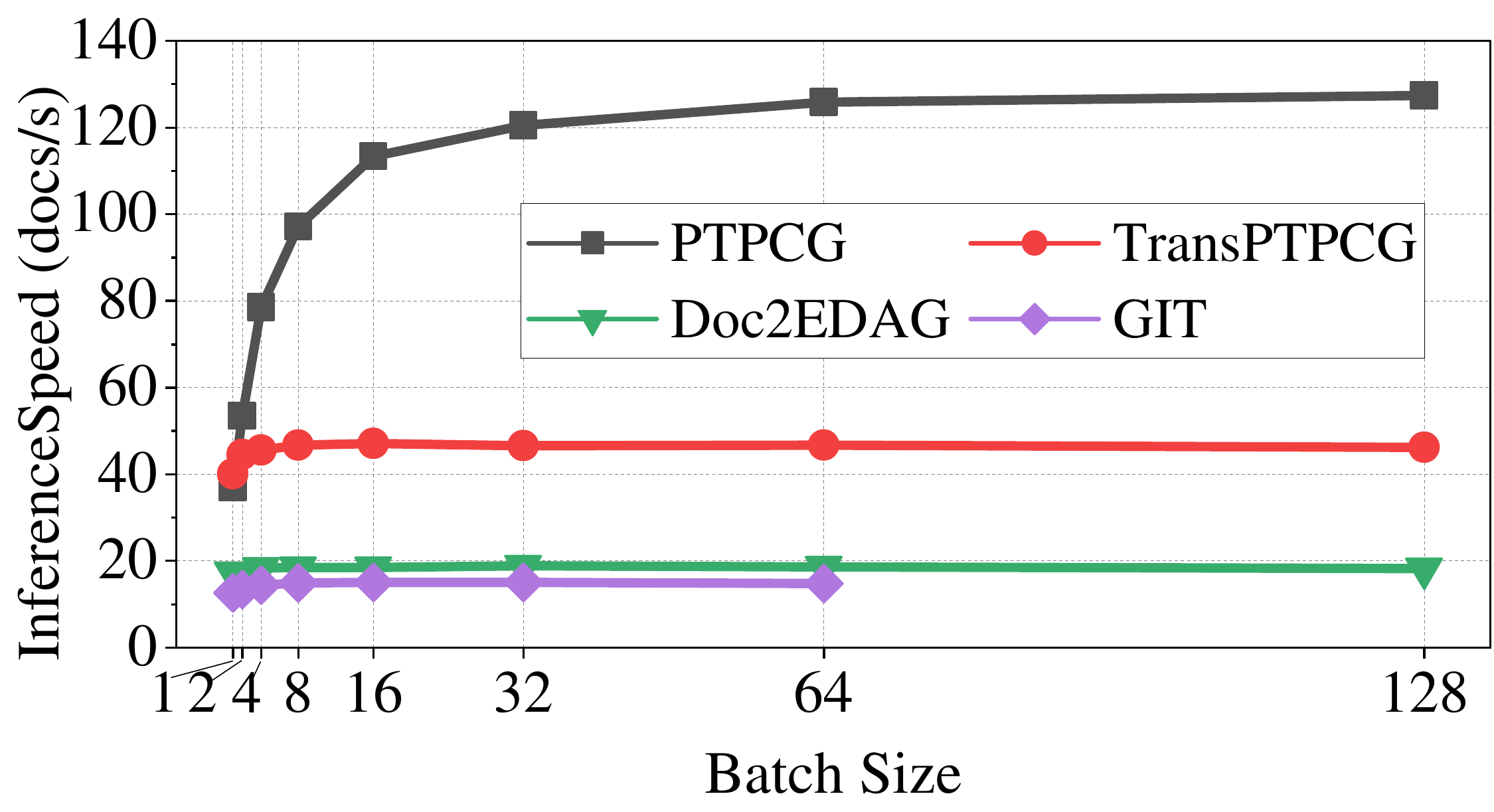}
    \caption{Inference speed comparison with baselines (left) and with different $|\mathcal{R}|$ (right) with 1 NVIDIA V100 GPU for all models.}
    \label{fig:inference_speed}
\end{figure}

According to the inference speed test in Figure~\ref{fig:inference_speed}, PTPCG is more scalable than other models.
With the growth of batch size, PTPCG becomes faster and finally stabilized near 125 documents per second, while Doc2EDAG and GIT are barely growing with batch size, peaking at 19 and 15 docs/s respectively.
PTPCG is up to 7.0 and 8.5 times faster compared to Doc2EDAG and GIT.
% As shown in the right of Figure~\ref{fig:inference_speed}, the speed does not decrease too much as $|\mathcal{R}|$ increases, refelecting the stability of PTPCG.
As an enhanced version of Doc2EDAG, GIT is 21.2\% slower than Doc2EDAG on average, and raises OOM error on a 32GB memory GPU when batch size is 128.
To further check the effect of different encoders, we substitute all BiLSTM layers into transformer encoders and create the TransPTPCG model with the same scale to Doc2EDAG parameters.
The results in Figure~\ref{fig:inference_speed} show that TransPTPCG is up to 2.5 times faster than Doc2EDAG, validating the advantage of non-autoregressive combination decoding.

\subsection{Pseudo Triggers as Supplements}

PTPCG is capable of handling with or without manually annotated triggers, and the automatically selected pseudo triggers can be supplements to enhance the performance.

As shown in Table~\ref{tab:dueefin_comparison}, we find the pseudo triggers could assist the trigger words to boost the importance score from 62.9 to 93.7, and the results show that this helps identify combinations and bring 0.7\% improvement in offline dev evaluation and 0.8\% improvement in the online test set.

\begin{table}[b]
    \centering
    \small
    \begin{tabular}{cccccccc}
        \toprule
        Pseudo & \multirow{2}{*}{Impt.} & \multicolumn{3}{c}{Dev} & \multicolumn{3}{c}{Online Test} \\
        \cmidrule(lr){3-5}\cmidrule(lr){6-8}
        Trigger&& P & R & F1 & P & R & F1 \\
        \midrule
        % $\times$ & 62.9 & \textbf{73.5} & 59.4 & 65.7 & \textbf{67.0} & 50.1 & 57.3 \\
        % \checkmark & 93.7 & 68.8 & \textbf{64.2} & \textbf{66.4} & 62.0 & \textbf{54.8} & \textbf{58.1} \\
        $\times$ & 62.9 & \textbf{76.5} & 56.8 & 65.2 & \textbf{69.8} & 48.5 & 57.3 \\
        \checkmark & \textbf{93.7} & 70.5 & \textbf{62.3} & \textbf{66.2} & 63.6 & \textbf{53.4} & \textbf{58.1} \\
        \bottomrule
    \end{tabular}
    \caption{PTPCG results on DuEE-fin w/ annotated triggers. Impt denotes the importance score.}
    \label{tab:dueefin_comparison}
\end{table}

To further validate the effectiveness of importance scores for pseudo trigger selection, we select groups of pseudo triggers with the middle and the lowest importance scores instead of the highest ones and analyze the results in Table~\ref{tab:importance_validation}.
The results show that there is a positive correlation between the importance and overall scores.
Nevertheless, the highest importance score (88.3\%) is not equal to 100.0\%, which may limit the upper bound of decoding.
We will explain more about error analysis in Section~\ref{sec:err_analyze}.

\begin{table}[t]
    \centering
    \small
    \begin{tabular}{cccccccc}
        \toprule
        \multirow{2}{*}{Impt.} & \multicolumn{3}{c}{ChFinAnn} & \multirow{2}{*}{Impt.} & \multicolumn{3}{c}{DuEE-fin w/o Tgg}\\
        \cmidrule(lr){2-4}\cmidrule(lr){6-8}
        & P & R & F1 & & P & R & F1 \\
        \cmidrule(lr){1-4}\cmidrule(lr){5-8}
        88.3 & 83.7 & \textbf{75.4} & \textbf{79.4} & 83.8 & 66.7 & \textbf{54.6} & \textbf{60.0} \\
        62.5 & 88.7 & 60.9 & 72.2 & 37.0 & 65.5 & 49.6 & 56.4 \\
        22.4 & \textbf{90.5} & 59.5 & 71.8 & 15.0 & \textbf{74.2} & 45.3 & 56.3 \\
        \bottomrule
    \end{tabular}
    \caption{PTPCG results with different importance scores.}
    \label{tab:importance_validation}
\end{table}

\subsection{Error Analysis}\label{sec:err_analyze}

Although pruned complete graph structure is efficient for training and inference, it is not perfect in similarity calculation (adjacent matrix prediction) and combination decoding.
Here we analyze the upper bounds of the combination decoding algorithm and discuss the future directions.

Results in Table~\ref{tab:upper_bound} show that models with the lower number of pseudo triggers are better.
However, models with more pseudo triggers have higher upper bounds for decoding and have higher importance scores.
Why models with more pseudo triggers have greater importance but still result in lower performance?
Results in Table~\ref{tab:importance_validation} and \ref{tab:upper_bound} may answer this question.
Models with the same $|\mathcal{R}|$ has a strong correlation that higher importance brings higher metric scores which validates the effectiveness of the pseudo trigger selection strategy based on importance scores.
However, more pseudo triggers bring more links to connect, and the model may be not robust enough for predicting each connection correctly, leading to an adjacent accuracy decline.

Overall, it is a trade-off that more pseudo triggers improves the upper bounds and reduces combination error rates, but also brings new challenges to recover connections between entities.
We believe it is the future direction to improve the similarity calculation and adjacent matrix prediction.

\begin{table}[bth]
    \centering
    \small
    \begin{tabular}{crrrrrrr}
        \toprule
        $|\mathcal{R}|$ & Impt. & SE & ME & TotE & \#links & \makecell{Adj\\Acc.} & F1 \\
        \midrule
        1 & 88.3 & 5.0 & 37.5 & 14.6 & 10,502 & 65.8 & 79.4 \\
        2 & 95.7 & 1.0 & 20.4 & 6.7 & 23,847 & 59.1 & 77.7 \\
        3 & 97.2 & 0.9 & 18.0 & 5.9 & 55,961 & 56.7 & 74.9 \\
        4 & 97.6 & 0.5 & 16.9 & 5.3 & 75,334 & 58.2 & 74.0 \\
        5 & 97.8 & 0.4 & 13.9 & 4.4 & 88,752 & 59.5 & 73.1 \\
        all & 97.8 & 0.2 & 13.4 & 4.1 & 140,989 & 60.1 & 69.5 \\
        \bottomrule
    \end{tabular}
    \caption{
        Analysis of combination decoding errors on ChFinAnn.
        SE, ME, TotE are error upper bounds of single-record documents, multiple-record documents, and overall error respectively.
        \#links is the number of connected links among all graphs in the test set.
        Adj Acc is the accuracy of adjacent matrix prediction given golden entities.
        F1 is micro-averaged.
    }
    \label{tab:upper_bound}
\end{table}

%% file: related_work.tex
\section{Related Work}

Fully annotated datasets are usually small in scale \cite{rams,wikievent}.
DS-constructed datasets are large in scale, but they can hardly match triggers to records, so triggers are likely to be absent.
Unlike trigger-based methods~\cite{ed3c,eeqa}, \citet{ed-wo-tgg} argue that event types can be detected without triggers.
\citet{yang-etal-2018-dcfee} extract event records using entities in a window of key sentences.
It is efficient but misses lots of information from other sentences.
To fully utilize all entities in the whole document, Doc2EDAG \cite{zheng-etal-2019-doc2edag} formulates the argument combination to be a directed acyclic graph (DAG), making great progress in DEE.
GIT \cite{xu-etal-2021-git} is a variant of Doc2EDAG where a graph neural network is added to help entity encoding and further exploit the global memory mechanism during decoding.
DAG extracts combinations in an auto-regressive way, which is very time consuming and needs huge space in global memory module to store all previous paths.
Besides, Doc2EDAG and GIT are both large models, where Doc2EDAG and GIT use 12 and 16 layers of transformer encoders.
To train such models, a minimum of four 32GB GPUs would have to be run for almost a week.
\citet{scdee} utilize BERT~\cite{bert} as sentence representations and exploit relations between sentences via graph neural networks~\cite{gat} to help identify fixed number of combinations.
To speed up DEE, we propose a novel non-autoregressive decoding strategy and boost the training \& inference speed with competitive results compared with the previous works.

%% file: conclusion.tex
\section{Conclusion}

Pursuing fast and general document-level event extraction (DEE), we propose a non-autoregressive model named as PTPCG.
For DEE without triggers, we first select a group of pseudo triggers to build pruned complete graphs, and then train a lightweight model to extract all possible combinations, which enables non-autoregressive decoding and is up to 8.5x faster compared to SOTA models.
For DEE with annotated triggers, pseudo triggers also show the power to make improvement and are even better than annotated trigger-only method.
In summary, our model costs less resources yet achieves better or comparable results, compared with the previous systems.

Although PTPCG is not perfect in adjacent matrix prediction, we believe it has the ability to combine different DEE tasks and needs more explorations.

%% file: acknowledgement.tex
\section*{Acknowledgments}

This work was supported by the National Natural Science Foundation of China (Grant No. 61936010 and 61876115), the Priority Academic Program Development of Jiangsu Higher Education Institutions, and the joint research project of Huawei Cloud and Soochow University.
We would also like to thank the anonymous reviewers for their valuable comments.